\documentclass[runningheads]{llncs}
\usepackage{graphicx}
\usepackage{comment}
\usepackage{amsmath,amssymb} 
\usepackage{color}
\usepackage{multirow}
\usepackage[caption=false]{subfig}
\usepackage[misc]{ifsym}
\usepackage[width=122mm,left=12mm,paperwidth=146mm,height=193mm,top=12mm,paperheight=217mm]{geometry}

\begin{document}
\pagestyle{headings}
\mainmatter

\title{Unsupervised Sketch-to-Photo Synthesis} %

\titlerunning{ }

\author{Runtao Liu\inst{1*}
\and
Qian Yu\inst{2*(\textrm{\Letter})}
\and
Stella Yu\inst{3}
}
\authorrunning{ }
\institute{
Peking University\\
\email{runtao219@gmail.com}\\
\and
Beihang University / UC Berkeley\\
\email{qianyu@buaa.edu.cn}
\and
UC Berkeley / ICSI\\
\email{stellayu@berkeley.edu}
}

\def\fig#1{Fig.\ref{fig:#1}}

\maketitle

\begin{abstract}
Humans can envision a realistic photo given a free-hand sketch that is not only spatially imprecise and geometrically distorted but also without colors and visual details.  We study unsupervised sketch-to-photo synthesis for the first time, learning from \textit{unpaired} sketch-photo data where the target photo for a sketch is unknown during training. Existing works only deal with style change or spatial deformation alone, synthesizing photos from edge-aligned line drawings or transforming shapes within the same modality, e.g., color images.
Our key insight is to decompose unsupervised sketch-to-photo synthesis into a two-stage translation task:  First shape translation from sketches to grayscale photos and then content enrichment from grayscale to color photos. We also incorporate a self-supervised denoising objective and an attention module to handle abstraction and style variations that are inherent and specific to sketches.
Our synthesis is sketch-faithful and photo-realistic to enable sketch-based image retrieval in practice. An exciting corollary product is a universal and promising sketch generator that captures human visual perception beyond the edge map of a photo.

\end{abstract}
\def\figtask#1{
\begin{figure}[#1]\vspace{-10pt}
\centering
\includegraphics[clip,width=0.98\textwidth]{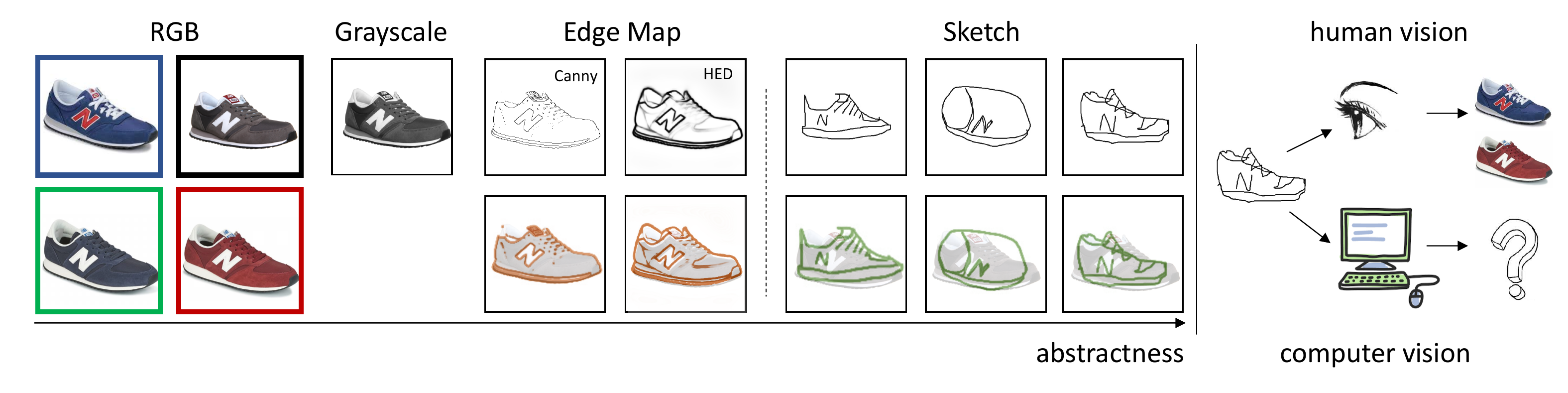}
\caption{%
The challenges of our unsupervised sketch-to-photo synthesis task. 
{\bf Left}: A single object could have multiple color realizations, but a common grayscale version. Edge maps extracted by different methods such as Canny and HED detectors lack colorful details but align well with the original object.  Human free-hand sketch is a line abstraction with various deformations and drawing styles.  The bottom row of \textit{Edge Map} and \textit{Sketch} shows the lines overlaid on the grayscale photo. 
{\bf Right}: Human vision can imagine a realistic photo given a free-hand sketch. Our goal of this work is to provide computer vision such an ability.
}
\label{fig:task}
\end{figure}
}

\def\figflow#1{
\begin{figure}[#1]
\centering
\includegraphics[width=0.98\textwidth,clip]{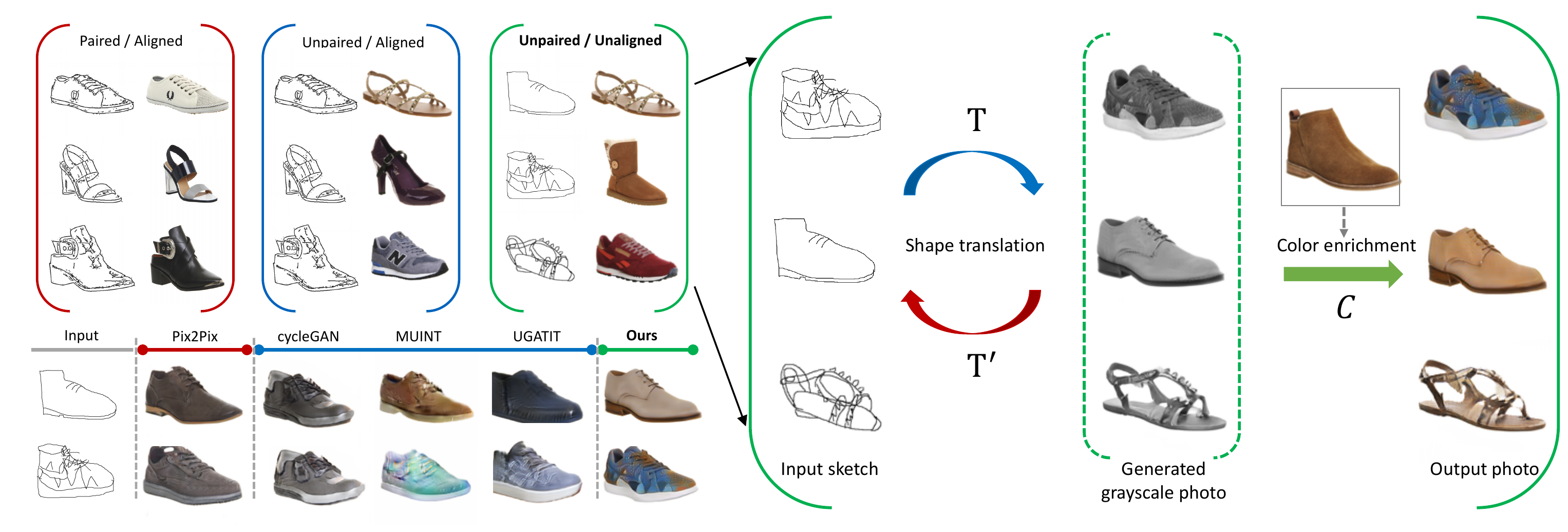}
\caption{\textbf{Left}: Comparison of sketch-to-photo different settings and results.
{\bf Top)} Three training scenarios on whether line-drawings and photos are provided as paired data and whether line-drawings are spatially aligned with the photos.  Edges extracted from photos are aligned, sketches are not.
{\bf Bottom)} Comparison of synthesis results.  Ours are superior to unsupervised edgemap-to-photo methods (cycleGAN \cite{zhu2017unpaired}, MUINT \cite{huang2018multimodal}, UGATIT \cite{UGATIT})
and even supervised methods (Pix2Pix \cite{isola2017image})  trained on paired data.  
\textbf{Right}: Our unsupervised sketch-to-photo synthesis model has two separate stages handling spatial deformation and colorful content fill-in challenges respectively.  {\bf 1)} Shape translation learns to synthesize a grayscale photo given a sketch, from unpaired sketch set and photo set.   {\bf 2)} Content enrichment learns to fill the grayscale photo with colorful details given an optional reference image.}
\label{fig:flow}
\end{figure}
}

\section{Introduction}

\figtask{tp}

Human free-hand sketch, {\it sketch} for short, is an intuitive and powerful visual expression (\fig{task}).   There is research on sketch recognition \cite{eitz2010evaluation,yu2015sketch}, sketch parsing \cite{yi2013icip,qi2015grouping}, and sketch-based image or video retrieval \cite{yu2016sketch,sangkloy2016sketchy,liu2017deep}. 
Here we study how to imagine a realistic photo given a sketch that is spatially imprecise and missing colors and details, by learning {\it unsupervisedly} from unpaired sketches and photos.

Sketch-to-photo synthesis is challenging for two reasons.
{\bf 1)} Sketch and photo are misaligned in shape since sketches commonly drawn by amateurs have large spatial and geometrical distortion.  Therefore, translating a sketch to a photo requires rectifying deformation.
{\bf 2)} Sketches are color-less and lacking visual details.  Drawn in black strokes on white paper, sketches outline mostly object boundaries and characteristic interior markings. To synthesize a photo, shading and colorful textures must be filled in properly.  

It is not trivial to rectify shape distortion, as line strokes are only suggestive of the actual shape and locations, and the extent of shape fidelity varies widely between individuals.  In \fig{task}, the three sketches for the same shoe are widely different, both globally (e.g., ratio) and locally (e.g., stroke style).  
It is not trivial to add visual details either.
Since a sketch could have multiple colorful realizations, any synthesized output must be both realistic and diverse.

Existing works thus focus on either shape or color translation alone (\fig{flow} Left). 
{\bf 1)}
Image synthesis that deals with shape transfiguration tends to stay in the same visual domain,  e.g. changing a picture of a dog to that of a cat \cite{liu2017unsupervised,huang2018multimodal}, where 
 visual details are comparable in the color image.
{\bf 2)}
Sketches are a special case of {\it line-drawings}, and the most studied case of line-drawings in computer vision is the edge map extracted automatically from a photo.  Such an edge map based lines-to-photo synthesis task does not have sketches' spatial deformation problem, and realistic images can be synthesized with  \cite{isola2017image,xian2018texturegan} or without \cite{zhu2017unpaired} paired data. 
We will show that existing methods fail in sketch-to-photo synthesis when both shape and color translations are needed simultaneously. 

Our key insight is to decompose this task into two separate translations.  Our two-stage model performs first geometrical shape translation in grayscale and then detailed content fill-in in color (\fig{flow} Right). 
{\bf 1)} The shape translation stage learns to synthesize a grayscale photo given a sketch, from unpaired sketch set and photo set.  Geometrical distortions are eliminated at this step.
{\bf 2)} The content enrichment stage learns to fill the grayscale with colorful details, including missing textures and shading, given an {\it optional} reference image. 

\figflow{h}

In order to handle abstraction and drawing style variations at Stage 1, we introduce a self-supervised learning objective and apply it to noise sketch compositions.  Additionally, we incorporate an attention module to help the model learn to ignore distractions. At Stage 2, a content enrichment network is designed to work with or without reference images.  This capability is enabled by a mixed training strategy.  Our model can thus produce diverse outputs.

Our model links sketches to photos and is directly applicable to sketch-based photo retrieval.  Another exciting corollary result from our model is that we can also synthesize a sketch given a photo, even from unseen semantic categories.  Strokes in a sketch capture information beyond edge maps which are defined primarily on intensity contrast and object exterior boundaries.  These automatic sketch results could lead to more advanced computer vision capabilities and serve
as powerful human-user interaction devices.

To summarize, our work makes the following major contributions. {\bf 1)} We propose the first two-stage unsupervised model that can generate diverse, sketch-faithful, and photo-realistic images from a single free-hand sketch. {\bf 2)} We introduce a self-supervised learning objective and an attention module to handle abstraction and style variations in sketches. {\bf 3)} Our work not only benefits sketch-based image retrieval but also delivers an automatic sketcher that captures human visual perception beyond the edge map of a photo.

\section{Related Works}

\noindent\textbf{Sketch-based image synthesis.} Much progress has been made on sketch recognition \cite{eitz2012hdhso,yu2015sketch,yu2017sketch} and sketch-based image retrieval \cite{eitz2011sbir,rui2010gfd,yi2014bmvc,yu2016sketch,sangkloy2016sketchy,liu2017deep}.  However, sketch-based image synthesis is still under-explored.  
Prior to deep learning (DL), Sketch2Photo \cite{chen2009sketch2photo} and PhotoSketcher \cite{eitz2011photosketcher} compose a new photo from photos retrieved for a given sketch. Sketch is also used for photo editing  \cite{bau2019semantic,portenier2018faceshop,sangkloy2017scribbler,yu2018free}.  

The first DL-based free-hand sketch-to-photo synthesis is SketchyGAN \cite{chen2018sketchygan} , which trains an encoder-decoder model conditioned on the class label for sketch-photo pairs.  \cite{ghosh2019interactive} focuses on multi-class photo generation based on incomplete edges or sketches.  While it also adopts a two-stage strategy for shape completion and appearance synthesis, it relies on paired training data and does not address the shape deformation challenge.

Photo-to-sketch has also been studied \cite{pang2018deep,song2018learning}.  While it is not our focus, our model trained only on shoe images can generate realistic sketches from photos in other semantic categories.

\noindent\textbf{Generative adversarial networks (GAN).}  GAN has a generator (G) and a discriminator (D): G tries to fake instances that fool D and D tries to detect fakes from reals.  GAN is widely used for realistic image generation  \cite{mirza2014conditional,karras2017progressive} and  translation across image domains \cite{isola2017image,huang2018multimodal}.

Pix2Pix \cite{isola2017image} is a conditional GAN that maps source images to target images; it requires paired data during training.  CycleGAN \cite{zhu2017unpaired} uses a pair of GANs to map an image from the source domain to the target domain and then back to the source domain.  Imposing a consistency loss over such a cycle of mappings, it allows both models to be trained together on unpaired images in two different domains.
UNIT \cite{huang2018multimodal}  and MUNIT \cite{huang2018multimodal} are the latest variations of cycleGAN.  

None of these methods work well when the source and target images are spatially poor aligned (\fig{task}) and exist in different color spaces.

\begin{figure}[!bp]
\centering
\includegraphics[width=0.98\textwidth]{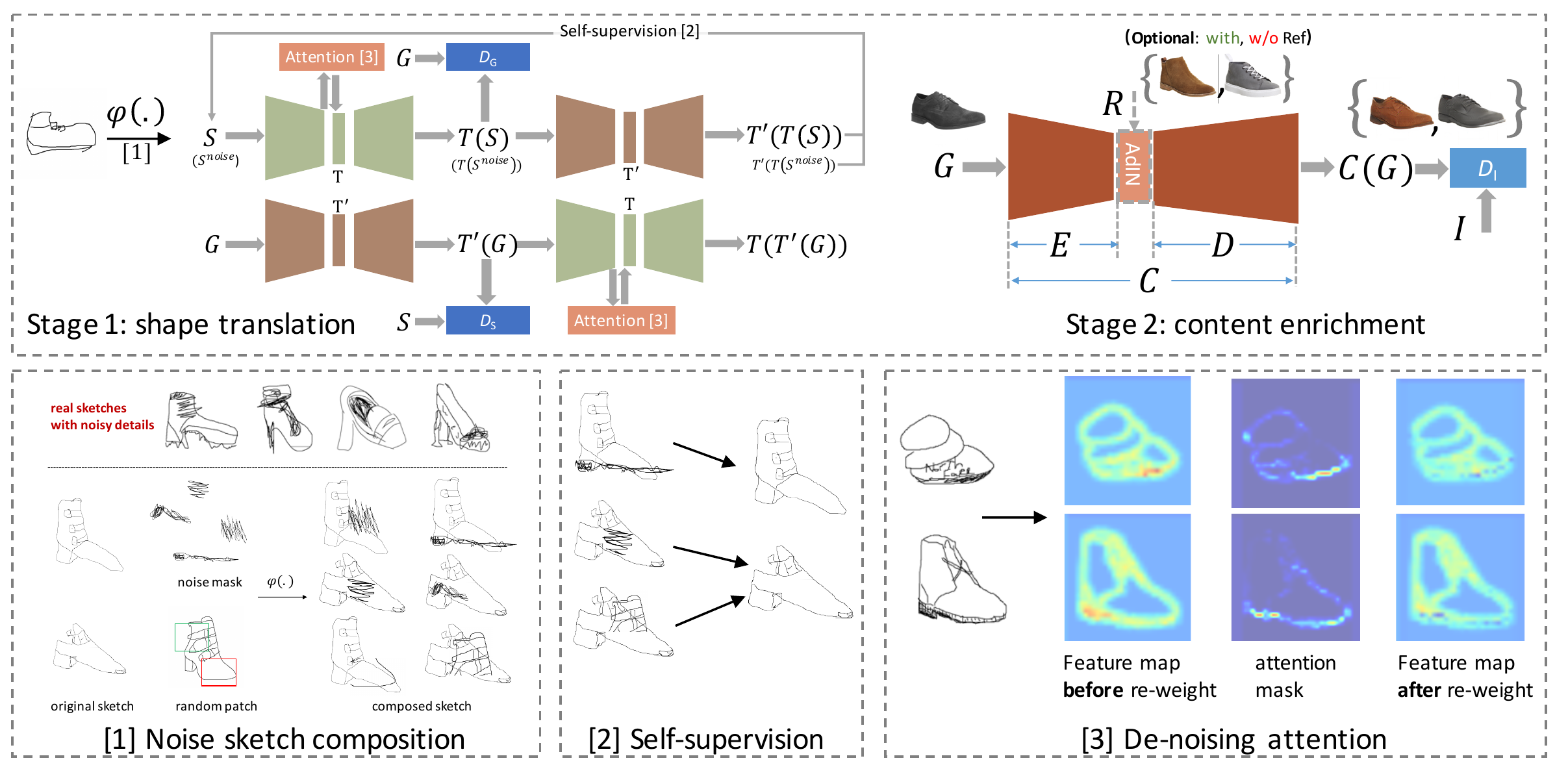}
\caption{\textbf{Top:} Our two-stage unsupervised model. \textbf{Bottom:} At the shape translation stage, we introduce noise sketch composition strategies, a self-supervised objective, and an attention module to tackle abstraction and style variations of sketches.}
\label{fig:architecture}
\end{figure}

\section{Two-Stage Sketch-to-Photo Synthesis}
\label{sec:method}

Compared to photos, sketches are spatially imprecise and colorless.  For sketch-to-photo synthesis, we deal with these two aspects at separate stages: We first translate a deformed sketch into a grayscale photo and then translate the grayscale into a color photo filled with missing details on texture and shading (\fig{architecture}).

Our unsupervised learning setting involves two data sets in the same semantic category such as shoes. Let there be $n$ sketches  $\{S_1,\ldots, S_n \}$,  $m$ color photos  $\{I_1,\ldots, I_m \}$ and their grayscale versions $\{G_1,\ldots, G_m \}$.

\subsection{Shape Translation: Sketch $S\to$ Grayscale $G$}
\label{sec:stage1}

\noindent\textbf{Overview.} We first learn to translate sketch $S$ into grayscale photo $G$. The goal is to rectify shape deformation in sketches.  We consider \textit{unpaired} sketch and photo images, since: {\bf 1)} Paired data are scarce and hard to collect; {\bf 2)} Given shape misalignment between sketches and photos, strong supervision imposed by paired data potentially confuses a model during training.

A pair of mappings, $T: S\xrightarrow{}G$ and  $T': G\xrightarrow{}S$ are learned with the constraint of cycle-consistency $S\approx T'(T(S))$ and $G\approx T(T'(G))$. Each has an encoder-decoder architecture. Similar to the model introduced in Zhu et al. \cite{zhu2017unpaired}, we train two domain discriminators $D_{G}$ and $D_{S}$.  $D_{G}$ tries to tease apart $G$ and $T(S)$, while $D_{S}$ teases apart $S$ and $T'(G)$ (Fig.~\ref{fig:architecture}(Top)).  $T(S)$ is the predicted grayscale to be fed into the subsequent content enrichment network. 

The input sketch may exhibit various levels of abstraction and different drawing styles.  In particular, sketches containing dense strokes or noisy details (Fig.~\ref{fig:architecture} Bottom left) cannot be handled well by a basic cycleGAN model.  

To deal with these variations, we introduce two strategies for the model to \textit{extract} only style-invariant information:
\textbf{1)} We compose additional noise sketches to enrich the dataset and introduce a self-supervised objective; \textbf{2)} We introduce an attention module to help detect distracting regions.

\noindent\textbf{Noise sketch composition.}  There are two kinds of noise sketches: {\it complex} sketches and \textit{distractive} sketches (Fig.~\ref{fig:architecture}(Bottom right)), denoted by $S^{noise}=\varphi(S)$, where $\varphi(.)$ represents composition.  We identify dense stroke patterns and construct a pool of noise masks.  We randomly sample from these masks and artificially generate \textit{complex} sketches by inserting these dense stroke patterns into original sketches.  We generate \textit{distractive} sketches by adding a random patch from a different sketch on an existing sketch.  The noise strokes and random patches are used to simulate irrelevant details in a sketch. We compose such noise sketches on-the-fly and feed them into the network with a fixed occurrence ratio.

\noindent\textbf{Self-Supervised objective.} We introduce a self-supervised objective to work with the synthesized noise sketches.  For a composed noise sketch, the reconstruction goal of our model is to reproduce the \textit{original clean} sketch:
\begin{equation}
 L_{ss}(T, T^\prime)= \left\|S-T^\prime \left(T(S^{noise})\right)\right\|_{1}
\label{eq:ss}
\end{equation}
This objective is different from the cycle-consistency loss used for other untouched sketches.  The new objective makes the model ignoring irrelevant strokes and putting more efforts on style-invariant strokes in the sketch.

\noindent\textbf{Ignore distractions with active attention.}  In addition to the above self-supervised training, we introduce an attention module to actively identify distracting stroke regions.  Since most areas of a sketch are blank, the activation of dense stroke regions is stronger than others.  Based on this intuition, we can locate distracting areas and \textit{suppress} the activation there accordingly. That is, the attention module generates an attention map $A$, and it is used to re-weight the feature representation of sketch $S$ (Eq. \ref{eq:feature}). $f(.)$ refers to the feature map and $\odot$ means element-wise multiplication.
\begin{align}
f_{\text{final}}(S)=(1-A)\odot f(S)
\label{eq:feature}
\end{align}
Unlike existing models using attention to highlight the region of interest, in our model the attended areas are weighted less. 

Our total objective for training a shape translation model is:
\begin{align*}
\min _{T, T^\prime} \max _{D_{G}, D_{S}} & \lambda_{1} (L_{adv}(T,D_{G};S,G) + L_{adv}(T^\prime,D_{S};G,S))  \\ + & \lambda_{2} L_{cycle}(T,T^\prime;S,G)+\lambda_{3} L_{identity}(T,T^\prime;S,G)+ L_{ss}(T,T^\prime;S^{noise}).
\label{eq:shape}
\end{align*}
We follow Zhu et al. \cite{zhu2017unpaired} to add an $L_{identity}$, which slightly improves the performance.  See the details of each loss in the Supplementary.

\subsection{Content Enrichment: Grayscale $G\to$ Color $I$}

The goal of content enrichment is to enrich the generated grayscale photo $G$ with missing details, learning a mapping $C$ that turns grayscale $G$ into color photo $I$.  
Since a color-less sketch could have many colorful realizations, many fill-in's are possible.  We thus model the task as a style transfer task and use an {\it optional} reference color image to guide the selection of a particular style.

We implement $C$ as an encoder-$E$ and decoder-$D$ network (Fig.~\ref{fig:architecture}(Top)).  Given a grayscale photo $G$ as the input, the model outputs a color photo.  The input and output images should be the same in CIE \textit{L}ab color space.  Therefore we use a self-supervised intensity loss (Eq.~\ref{eq:L2}) to train the model.
Additionally, a discriminator $D_{I}$ is trained to ensure the photo-realism of the output.

\begin{equation}
L_{it}(C) = \left\|G-Lab\left(C\left(G \right)\right)\right\|_{1}
\label{eq:L2}
\end{equation}

To improve the diversity of output, a conditional module is introduced to accept a reference image for guidance.  We follow AdaIN \cite{huang2017arbitrary} to inject style information by adjusting the statistics of feature map.  Specifically, the encoder $E$ encodes the input grayscale image $G$ and generates a feature map $\textbf{x}=E(G)$, then the mean and variance of \textbf{x} are adjusted by reference's feature map $\textbf{x}^{ref}=E(R)$.  The new feature map is $\textbf{x}^{new}=AdaIN(\textbf{x},\textbf{x}^{ref})$ (Eq.\ref{eq:adin}) and is sent to the decoder $D$ for rendering the final output.
\begin{align}
AdaIN(\textbf{x},\textbf{x}^{ref})&= \sigma(\textbf{x}^{ref})(\frac{\textbf{x}-\mu(\textbf{x})}{\sigma (\textbf{x})})+\mu(\textbf{x}^{ref})
\label{eq:adin}
\end{align}

Our model can work with or without reference images, in a \textit{single} network,  enabled by a mixed training strategy.  When there is no reference image, only intensity loss and adversarial loss are used while $\sigma(\textbf{x}^{ref})$ and $\mu(\textbf{x}^{ref})$ are set to 1 and 0 respectively; otherwise, a content loss and style loss are computed additionally.  The content loss is used to guarantee that the input and output images are consistent perceptually, whereas the style loss is to ensure the style of the output is aligned with that of the reference image. 
The total loss for training the content enrichment model is:
\begin{align}
\min _{C} \max _{D_{I}} & \lambda_{4} L_{adv}(C,D_{I};G,I)+ \lambda_{5} L_{it}(C)  +  \lambda_{6} L_{style}(C;G,R)+\lambda_{7} L_{cont}(C;G,R)
\label{eq:step2}
\end{align}
Network architectures and further details are provided in the Supplementary. 

\def\imw#1#2{\includegraphics[width=#2\linewidth]{#1}}
\def\imh#1#2{\includegraphics[height=#2\textheight]{#1}}
\def\imwh#1#2#3{\includegraphics[width=#2\linewidth,height=#3\textheight]{#1}}

\newcommand{\tb}[3]{\setlength{\tabcolsep}{#2mm}\begin{tabular}{#1}#3\end{tabular}}

\def\row#1#2#3#4{
\imw{figs/supp_2_sketch2photo/input/#4_#1.png}{0.11}&
\imw{figs/supp_2_sketch2photo/ours/#4_#1.png}{0.11}&
\imw{figs/supp_2_sketch2photo/ref1/#4_#1.png}{0.11}&
\imw{figs/supp_2_sketch2photo/input/#4_#2.png}{0.11}&
\imw{figs/supp_2_sketch2photo/ours/#4_#2.png}{0.11}&
\imw{figs/supp_2_sketch2photo/ref1/#4_#2.png}{0.11}&
\imw{figs/supp_2_sketch2photo/input/#4_#3.png}{0.11}&
\imw{figs/supp_2_sketch2photo/ours/#4_#3.png}{0.11}&
\imw{figs/supp_2_sketch2photo/ref1/#4_#3.png}{0.11}\\
}
\def\chairrow#1{
\imw{figs/baseline/input/chair#1.png}{0.12}& 
\imw{figs/baseline/pix2pix/chair#1.png}{0.12}&
\imw{figs/baseline/cyclegan/chair#1.png}{0.12}&
\imw{figs/baseline/munit/chair#1.png}{0.12}&
\imw{figs/baseline/ugatit/chair#1.png}{0.12}&
\imw{figs/baseline/ours/chair#1.png}{0.12}&
\imw{figs/baseline/ref1/chair#1.png}{0.12}&
\imw{figs/baseline/ref2/chair#1.png}{0.12}\\
}
\def\shoesrow#1{
\imw{figs/baseline/input/shoes_#1.png}{0.12}& 
\imw{figs/baseline/pix2pix/shoes_#1.png}{0.12}&
\imw{figs/baseline/cyclegan/shoes_#1.png}{0.12}&
\imw{figs/baseline/munit/shoes_#1.png}{0.12}&
\imw{figs/baseline/ugatit/shoes_#1.png}{0.12}&
\imw{figs/baseline/ours/shoes_#1.png}{0.12}&
\imw{figs/baseline/ref1/shoes_#1.png}{0.12}&
\imw{figs/baseline/ref2/shoes_#1.png}{0.12}\\
}

\def\figbaseline#1{
\begin{figure*}[#1]\centering
{\small
    \tb{@{}c|c|ccc|ccc@{}}{0.1}{
    \shoesrow{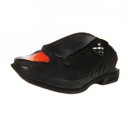}
    \shoesrow{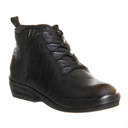}
    \shoesrow{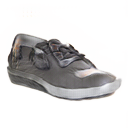}
    \chairrow{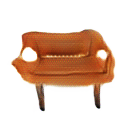}
    \chairrow{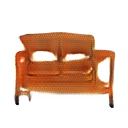}
    \chairrow{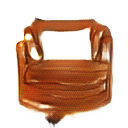}
    Input&
    Pix2Pix&
    cycleGAN&
    MUNIT&
    UGATIT&
    \textbf{Ours}&
    Ours\_ref1  &
    Ours\_ref2 \\
    }
}

\vspace{0.3cm}
{\small
    \tb{@{}ccc|ccc|ccc@{}}{0.05}{
        \row{3}{5}{6}{shoes_supp}
        \row{10}{4}{0}{shoes_supp}
        \row{5}{1}{2}{chairs_supp}
        \row{12}{7}{4}{chairs_supp}
        Input&
        Ours&
        Ours\_ref &
        Input&
        Ours&
        Ours\_ref &
        Input&
        Ours&
        Ours\_ref \\
    }
}\\
\caption{\small Our model can produce high-fidelity and diverse photos based on a sketch. \textbf{Top:} comparisons of our model with baselines. Most of these methods cannot handle this task well. While methods like UGATIT can generate realistic photos, but our results are more faithful to the input sketch, e.g., the three chair examples. \textbf{Bottom:} Synthesized results obtained by our model, with (3rd column) and without (2nd column) reference image. Note that our content enrichment model can work under both settings(with or without reference) with a \textit{single} network. Reference images are shown in the top right. 
\label{fig:baseline}
}
\end{figure*}
}

\section{Experimental Setup and Evaluation Metrics}

\noindent\textbf{Datasets.}
We train our model on two datasets, ShoeV2 and ChairV2 \cite{yu2016sketch}. ShoeV2 is the largest single-class sketch dataset with 6,648 sketches and 2,000 photos. ChairV2 has  1,297 sketches and 400 photos.  Each photo has at least 3 corresponding sketches drawn by different individuals.  Note that we do not use paired information during training. 

Compared with other existing sketch datasets such as QuickDraw \cite{ha2017neural}, Sketchy \cite{sangkloy2016sketchy}, and TU-Berlin \cite{eitz2012hdhso}, these two datasets not only contain sketch and photo images, but their sketches have more fine-grained details.  They pose a more challenging setting where the synthesized photos must reflect these details.

\noindent\textbf{Baselines.}
We choose 4 image translation baselines.
\begin{enumerate}
\setlength{\itemsep}{0pt}
\item \textbf{Pix2Pix} \cite{isola2017image} is a conditional generative model and requires paired images of two domains for training.  It serves as a supervised learning baseline.

\item \textbf{CycleGAN} \cite{zhu2017unpaired} is a bidirectional unsupervised image-to-image translation model. It is the first model to apply cycle-consistency in GAN-based image translation task and it allows a model to be trained on unpaired data. 

\item \textbf{MUNIT}\cite{huang2018multimodal} is also an unsupervised model with the target of generating multiple outputs given an input. It assumes that the representation of an image can be decomposed into a content code and a style code. The model learns these two codes simultaneously. 

\item \textbf{UGATIT} \cite{UGATIT} is an attention-based image translation model. The proposed attention module is designed to help the model focus on the domain-discriminative regions, which would assume more weights to improve the quality of the synthesized results. 

\end{enumerate}

\noindent\textbf{Training details.}
We train our shape translation network for 500 epochs on shoes (400 for chairs), and content enrichment network for 200 epochs. The initial learning rate is set to  0.0002, and the input image size is $128\times128$. We use Adam optimizer with batch size 1. Using a larger batch size can speed up the training process but lead to performance drop slightly.  Following the practice introduced in cycleGAN, we train the first 100 epochs at the same learning rate and then linearly decrease the rate to zero until the maximum epoch. During training shape translation network, we randomly compose \textit{complex} and \textit{distractive} sketches with the possibility of 0.2 and 0.3 respectively. The random patch size is $50\times50$. We implement the attention module as a two-layer convolutional network. The Softmax activation function is used to produce the attention mask. When training the content enrichment network, reference images are fed into the network with a small possibility of 0.2.

\noindent\textbf{Evaluation metrics.}
We use three metrics.
\begin{enumerate}
\setlength{\itemsep}{0pt}
\item \textbf{Fréchet Inception Distance} (FID).
{It measures the distance between generated samples and real samples according to the statistics of activation distributions in a pre-trained Inception-v3 pool3 layer.  It could evaluate quality and diversity simultaneously. Lower FID value indicates higher fidelity. }

\item \textbf{User study} (Quality). We ask users to evaluate the similarity and realism of results produced by different methods. Following  \cite{wang2018high}, we ask human subjects (4 individuals who know nothing about our work) to compare two generated photos and select the one which they think fits their imagination better for a given sketch. We sample 50 pairs for each comparison.

\item \textbf{Learned perceptual image patch similarity} (LPIPS).
It evaluates the distance between two images.  As in \cite{huang2018multimodal} and \cite{zhu2017unpaired}, we utilize this metric to evaluate the \textit{diversity} of the outputs generated by different methods. 
\end{enumerate}

\begin{table}[t]
\centering
\caption{Benchmarks on ShoeV2/ChairV2. `$*$' indicates paired data for training. }
\setlength{\tabcolsep}{5pt}
\begin{tabular}[t]{c|ccc||ccc}
\hline
& & ShoeV2 &  &  & ChairV2\\
\hline
Model & FID $\downarrow$ & Quality $\uparrow$ & LPIPS $\uparrow$ & FID $\downarrow$ & Quality $\uparrow$ & LPIPS $\uparrow$ \\
\hline
Pix2Pix$^{*}$ & 65.09 & 27.0 & 0.071 & 177.79 & 13.0 & 0.096 \\
CycleGAN      & 79.35 & 12.0 & 0.0 & 124.96 & 20.0 & 0.0\\
MUNIT         & 92.21 & 14.5 & \textbf{0.248} & 168.81 & 6.5 & \textbf{0.264} \\
UGATIT        & 76.89 & 21.5 & 0.0 & 107.24 & 19.5 & 0.0 \\
\hline
Ours          & \textbf{48.73} & \textbf{50.0} & 0.146 & \textbf{100.51} & \textbf{50.0} & 0.156 \\
\hline
\end{tabular}
\label{tab:FID}
\end{table}

\section{Experimental Results}

\subsection{Sketch-based Photo Synthesis}

\noindent{\bf Benchmarks in Table \ref{tab:FID}.}  {\bf 1)}  Our model outperforms all baselines in FID and user studies. Note that all baseline models have a one-stage architecture.
{\bf 2)} All models perform poorly on ChairV2, probably due to more shape variations but far fewer training data for chairs than for shoes (1:5).
{\bf 3)} Ours outperforms MUNIT by a large margin. This indicates that our task-level decomposition strategy, i.e., two-stage architecture, is more suitable for sketch-to-photo synthesis than feature-level decomposition. 
{\bf 5)} UGATIT ranks the second on each dataset. It is also an attention-based model, showing the effectiveness of attention in image translation tasks. 

\begin{figure}[t]
\centering
\includegraphics[width=0.98\textwidth]{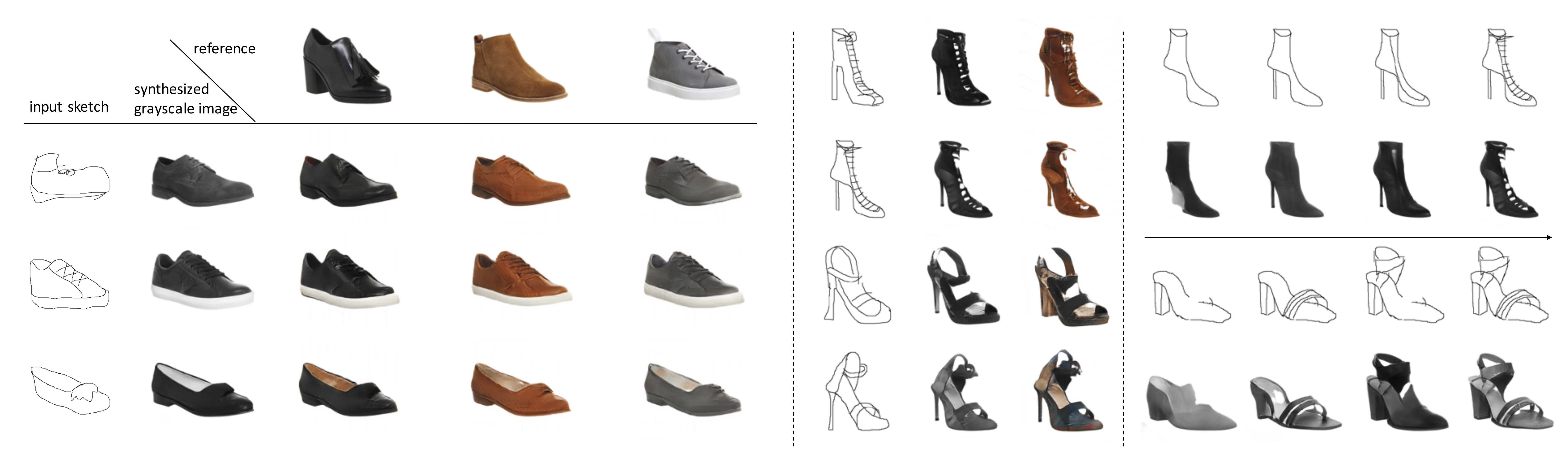}
\caption{\textbf{Left:} With different references, our model can produce diverse outputs. \textbf{Middle:} Given sketches of similar shoes drawn by different users, our model can capture their commonality as well as subtle distinctions and translate them into photos. Each row shows one example, including the input sketch, synthesized grayscale image, synthesized RGB photo. {\bf Right:} Our model even works for sketches at different completion stages, delivering realistic closely looking shoes.}
\label{fig:ref}
\end{figure}

\noindent{\bf Comparisons in \fig{baseline} and varieties in \fig{ref}(Left).}  Our results are more realistic and faithful to the input sketch (e.g., buckle and logo); our synthesis with different reference images produces varieties.

\figbaseline{tp}

\def\row#1#2{
\imw{figs/fig_acrossdomain_cycleABA/input/#1.png}{0.1}& 
\imw{figs/fig_acrossdomain_cycleABA/grayscale/#1.png}{0.1}&
\imw{figs/fig_acrossdomain_cycleABA/rgb/#1.png}{0.1}&
\imw{figs/fig_acrossdomain_cycleABA/with_ref/#1.png}{0.1}&
\hspace{0.8cm}
\imw{figs/fig_acrossdomain_cycleABA/a/#2.png}{0.1}&
\imw{figs/fig_acrossdomain_cycleABA/b/#2.png}{0.1}&
\imw{figs/fig_acrossdomain_cycleABA/c/#2.png}{0.1}&
\imw{figs/fig_acrossdomain_cycleABA/d/#2.png}{0.1}\\
}
\begin{figure}[ht]\centering
{\small
\tb{@{}cccccccc@{}}{0.05}{
\row{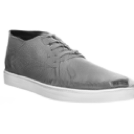}{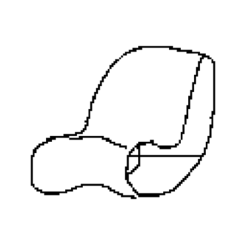}
\row{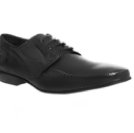}{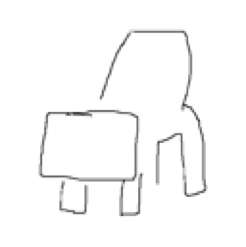}
\row{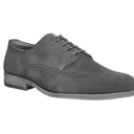}{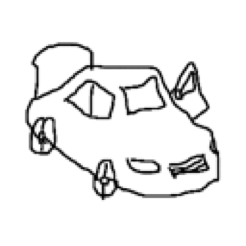}
\row{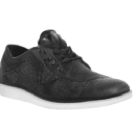}{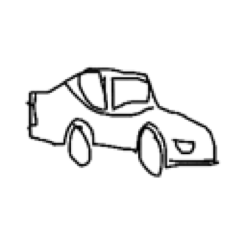}
Input&
Grayscale&
RGB&
With ref.&
\hspace{0.8cm}
(a)&
(b)&
(c)&
(d)\\
}
}
\caption{\textbf{Left:} Generalization across domains.  Column 1 are sketches from two unseen datasets, Sketchy and TU-Berlin.  Columns 2-4 are results from our model trained on ShoeV2. \textbf{Right:} Our shoe model can be used as a shoe detector and generator. It can generate a shoe photo based on a non-shoe sketch.  It can further turn the non-shoe sketch into a more shoe-like sketch. (a) Input sketch; (b) synthesized grayscale photo; (c) re-synthesized sketch; (d) Green \emph{(a)} overlaid over gray \emph{(c)}.}
\label{fig:robust}
\end{figure}

\noindent\textbf{Robustness and Sensitivity in Fig.~\ref{fig:ref}(Middle\&Right).}
We test our model trained on ShoeV2 under two settings: 1) sketches corresponding to the same photo, 2) sketches at different completed stages.
Given sketches of similar shoes drawn by different users, our model can capture their commonality as well as subtle distinctions and translate them into photos.  
Our synthesis model also works for sketches at different completion stages, obtained by removing strokes 
according to the stroke sequence available in ShoeV2.
Our model synthesizes realistic closely-looking shoes for partial sketches. 

\noindent\textbf{Generalization across domains in Fig.\ref{fig:robust}(Left)}.
When sketches are randomly sampled from different datasets such as TU-Berlin \cite{eitz2012hdhso} and Sketchy \cite{sangkloy2016sketchy}, which have greater shape deformation than ShoeV2, our model trained on ShoeV2 can still produce good results (more in the Supplementary).

\noindent\textbf{Sketches from novel categories in Fig.\ref{fig:robust}(Right)}. While we focus on a single category training, we nonetheless feed our model sketches from other categories.  When the model is trained on shoes, the shape translation network has learned to synthesize a grayscale shoe photo based on a \textit{shoe} sketch.
For a non-shoe sketch, our model translates it into a shoe-like photo. Some fine details in the sketch become a common component of a shoe. For example, a car becomes a trainer while the front window becomes part of a shoelace.  The superimposition of the input sketch and the re-\textit{shoe}-synthesized sketch reveals which lines are chosen by our model and how it modifies the lines for re-synthesis.

\subsection{Ablation Study}
\label{sec:ablation}

\noindent\textbf{Two-stage architecture.} Two-stage architecture is the key to the success of our model. This strategy can be easily adapted by other models such as cycleGAN.  Table \ref{tab:twostep} compares the performance of the original cycleGAN and its two-stage version\footnote{In the two-stage version, cycleGAN is used \textit{only} for shape translation while the content enrichment network is the same as ours.}. The two-stage version outperforms the original cycleGAN by 27.55 (on ShoeV2) and 68.33 (on ChairV2), indicating the significant benefits brought by this architectural design. 

\begin{table}[t]
\centering
\caption{Comparison of different architecture designs.}
\begin{tabular}[t]{c|cccc}
\hline
FID $\downarrow$ & cycleGAN(1-stage) & cycleGAN(2-stage) & Edge Map & Grayscale(Ours) \\
\hline
ShoeV2 &  79.35 & 51.80 & 96.58 & \textbf{48.73}\\
\hline
ChairV2 & 177.79 & 109.46 & 236.38 & \textbf{100.51} \\
\hline
\end{tabular}
\label{tab:twostep}
\end{table}

\def\row#1#2{
\imw{figs/fig_edge_pix/input/fig8_#1.png}{0.08}& 
\imw{figs/fig_edge_pix/edge/fig8_#1.png}{0.08}&
\imw{figs/fig_edge_pix/rgb/fig8_#1.png}{0.08}&
\imw{figs/fig_edge_pix/ours/fig8_#1.png}{0.08}&
\hspace{0.8cm}
\imw{figs/fig_att/input/#2.png}{0.08}& 
\imw{figs/fig_att/att_mask/#2.png}{0.08}&
\imw{figs/fig_att/wo_att/#2.png}{0.08}&
\imw{figs/fig_att/with_att/#2.png}{0.08}&
\imw{figs/fig_att/full/#2.png}{0.08}&
\imw{figs/fig_att/ugatit/#2.png}{0.08}\\
}
\begin{figure}[t]
\centering
\subfloat{\small
    \tb{@{}cccccccccc@{}}{0.05}{
    \row{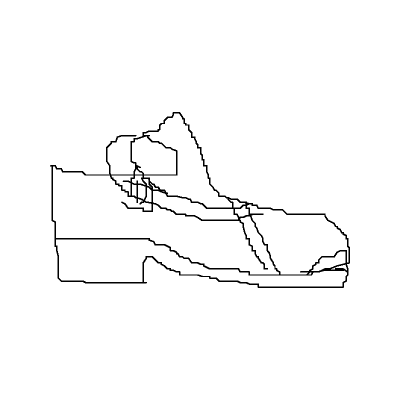}{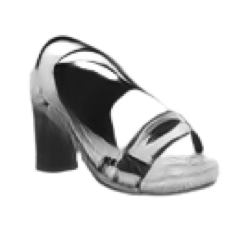}
    \row{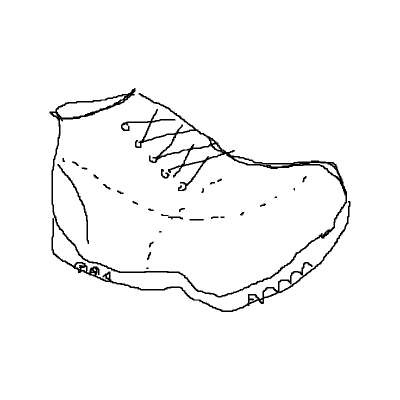}{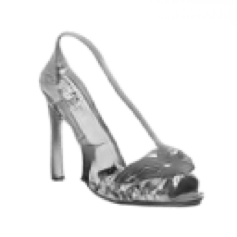}
    (a)&
    (b)&
    (c)&
    (d)&
    \hspace{0.6cm}
    (a)&
    (b)&
    (c)&
    (d)&
    (e)&
    (f)\\
    }
}

\caption{\textbf{Left:} Synthesized results when the edge map is used as the intermediate goal instead of the grayscale photo. (a) Input sketch; (b) Synthesized edge map, (c) Synthesized RGB photo using the edge map; (d) Synthesized RGB photo using grayscale (Ours). \textbf{Right:} Our model can successfully deal with noise sketches, which are not well handled by another attention-based model, UGATIT. For an input sketch (a), our model produce an attention mask (b); (c) and (d) are grayscale images produced by vanilla and our model. (e) and (f) compare ours with the result of UGATIT.}
\label{fig:proxy}
\end{figure}

\def\prow#1#2#3#4{
\imw{figs/fig_edge_pix/pix_input/figpix_#1.png}{0.08}&
\imw{figs/fig_edge_pix/pix_gray/figpix_#1.png}{0.08}&
\imw{figs/fig_edge_pix/ours_gray/figpix_#1.png}{0.08}&
\imw{figs/fig_edge_pix/pix_input/figpix_#2.png}{0.08}&
\imw{figs/fig_edge_pix/pix_gray/figpix_#2.png}{0.08}&
\imw{figs/fig_edge_pix/ours_gray/figpix_#2.png}{0.08}&
\imw{figs/fig_edge_pix/pix_input/figpix_#3.png}{0.08}&
\imw{figs/fig_edge_pix/pix_gray/figpix_#3.png}{0.08}&
\imw{figs/fig_edge_pix/ours_gray/figpix_#3.png}{0.08}&
\imw{figs/fig_edge_pix/pix_input/figpix_#4.png}{0.08}&
\imw{figs/fig_edge_pix/pix_gray/figpix_#4.png}{0.08}&
\imw{figs/fig_edge_pix/ours_gray/figpix_#4.png}{0.08}\\
}
\begin{figure}[t]
\centering
\subfloat{\small
    \tb{@{}ccc|ccc|ccc|ccc@{}}{0.05}{
    \prow{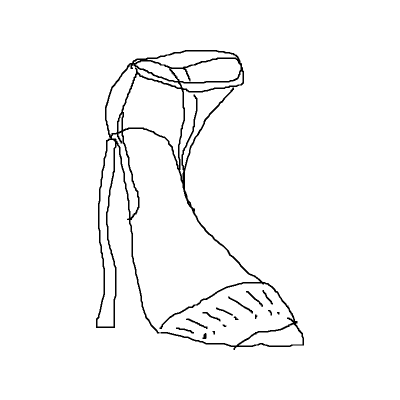}{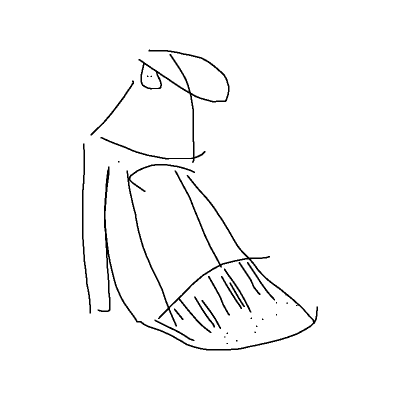}{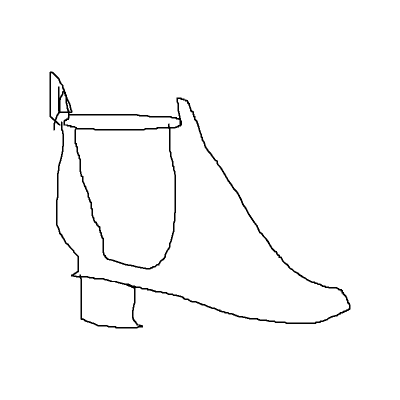}{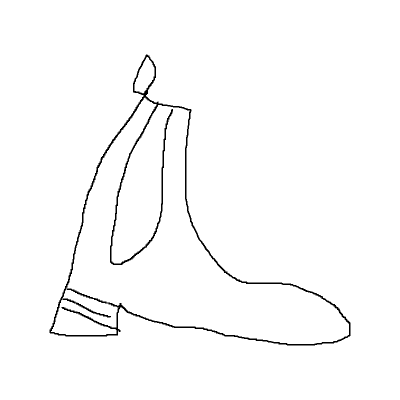}

    }
}
\caption{Comparisons of paired and unpaired training for shape translation. There are four examples. For each example, the 1st one is the input sketch, the 2nd and the 3rd are grayscale images synthesized by Pix2Pix and our model respectively.  Note that for each example, although the input sketches are different visually, Pix2Pix produces a similar-looking grayscale image.  Our results are more faithful to the sketch.}
\label{fig:pair}
\end{figure}

\begin{table}[t]
\centering
\caption{Contribution of each proposed component. The FID scores are obtained based on the results of \textit{shape translation stage}. \textit{SS.}: self-supervised objective, \textit{Att.}: attention.}
\setlength{\tabcolsep}{12pt}
\begin{tabular}[t]{c|cccccc}
\hline
FID $\downarrow$ & Pix2Pix & Vanila & w/o SS. & w/o Att. & Ours\\
\hline
ShoeV2 & 75.84 & 48.30 & 46.88 & 47.0 & \textbf{46.46}\\
\hline
ChairV2 & 164.01 & 104.0 & 93.33 & 92.03 & \textbf{90.87} \\
\hline
\end{tabular}
\label{tab:component}
\end{table}

\noindent\textbf{Edge map vs. grayscale as the intermediate goal.}  We choose \textit{grayscale} as our intermediate goal of translation.  As shown in Fig.~\ref{fig:task}, \textit{edge maps} could be an alternative since it does not have shape deformation either.  We can first translate sketch to an edge map, and then fill the edge map with colors and textures to get the final result.

Table \ref{tab:twostep} and Fig.~\ref{fig:proxy} show that using the edge map is worse than using the grayscale.  Our explanations are: \textbf{1)} Grayscale images have more visual information so that more learning signals are available when training the shape translation model; \textbf{2)} Content enrichment is easier for the grayscale as they are closer to color photos than edge maps.  The grayscale is also easier to obtain in practice.

\noindent\textbf{Deal with abstraction and style variations.} 
We have discussed the problem encountered during shape translation in Section \ref{sec:stage1}, and further introduced 1) a self-supervised objective along with noise sketch composition strategies and 2) an attention module to handle the problem. 
Table \ref{tab:component} compares FID achieved at the first stage by different variants.  Our full model can tackle the problem better than the vanila model, and each component contributes to the improved performance.  Figure \ref{fig:proxy} shows two examples and compares the results of UGATIT.

\begin{table}[h]
\centering
\caption{Exclude the effect of paired data. Although the paired information is not used during training, they indeed exist in ShoeV2.  We compose a new dataset where pairing does not exist, and use this dataset to train the model again. The results are obtained on the same test set.}
\setlength{\tabcolsep}{10pt}
\begin{tabular}[t]{c|c|c|c}
\hline
Dataset & Paired Exist? & Use Pair Info. & FID $\downarrow$\\
\hline
ShoeV2 & Yes & No & \textbf{48.7}\\
\hline
UT Zappos50K & No & No & \textbf{48.6} \\
\hline
\end{tabular}
\label{tab:paired}
\end{table}

\noindent\textbf{Paired vs. unpaired training.} We train a Pix2Pix model for shape translation to see if paired information helps.  As shown in Table \ref{tab:component} (\textit{Pix2Pix}) and Fig.~\ref{fig:pair}, its performance is much worse than ours (FID: 75.84 vs. 46.46 on ShoeV2 and 164.01 vs. 90.87 on ChairV2), most likely caused by the shape misalignment between sketches and grayscale images.

\noindent\textbf{Exclude the effect of paired information.}  Although pairing information is not used during training, they do exist in ShoeV2.  To eliminate any potential pairing facilitation, we train another model on a composed dataset, created by merging all the sketches of ShoeV2 and 9,995 photos of UT Zappos50K \cite{yu2014fine}. These photos are collected from a different source than ShoeV2.
We train this model in the same setting. In Table \ref{tab:paired}, we can see this model achieves similar performance with the one trained on ShoeV2, indicating the effectiveness of our approach for learning the task from entirely \textbf{unpaired} data.

\def\row#1#2#3#4#5{
    \imw{figs/fig_photo2sketch/shoe_photo/#1.png}{0.1}&
    \imw{figs/fig_photo2sketch/shoe_sketch/#1.png}{0.1}&
    \imw{figs/fig_photo2sketch/shoe_photo/#2.png}{0.1}&
    \imw{figs/fig_photo2sketch/shoe_sketch/#2.png}{0.1}&
    \imw{figs/fig_photo2sketch/shoe_photo/#3.png}{0.1}&
    \imw{figs/fig_photo2sketch/shoe_sketch/#3.png}{0.1}&
    \imw{figs/fig_photo2sketch/shoe_photo/#4.png}{0.1}&
    \imw{figs/fig_photo2sketch/shoe_sketch/#4.png}{0.1}&
    \imw{figs/fig_photo2sketch/shoe_photo/#5.png}{0.1}&
    \imw{figs/fig_photo2sketch/shoe_sketch/#5.png}{0.1}\\
}
\def\srow#1#2{
    \imw{figs/fig_photo2sketch/photo/#1.png}{0.1}&
    \imw{figs/fig_photo2sketch/canny/#1.png}{0.1}&
    \imw{figs/fig_photo2sketch/hed/#1.png}{0.1}&
    \imw{figs/fig_photo2sketch/photo_sketching/#1.png}{0.1}&
    \imw{figs/fig_photo2sketch/ours/#1.png}{0.1}&
    \imw{figs/fig_photo2sketch/photo/#2.png}{0.1}&
    \imw{figs/fig_photo2sketch/canny/#2.png}{0.1}&
    \imw{figs/fig_photo2sketch/hed/#2.png}{0.1}&
    \imw{figs/fig_photo2sketch/photo_sketching/#2.png}{0.1}&
    \imw{figs/fig_photo2sketch/ours/#2.png}{0.1}\\
}
\begin{figure}[t]
\centering
\subfloat{\small
    \tb{@{}cccccccccc@{}}{0.05}{
        \row{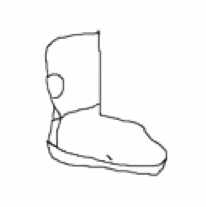}{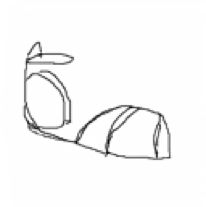}{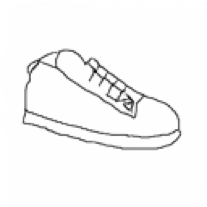}{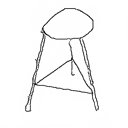}{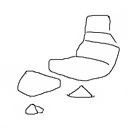}
        \row{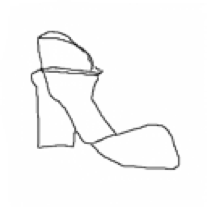}{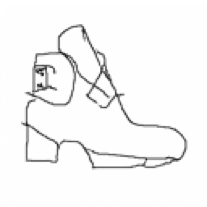}{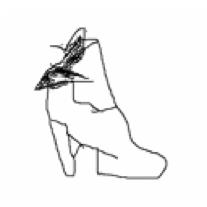}{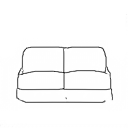}{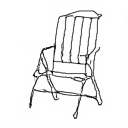}
    }
}

\vspace{-0.2cm}
\rule{12cm}{0.4pt}
\subfloat{\small
    \tb{@{}ccccc|ccccc@{}}{0.05}{
        \srow{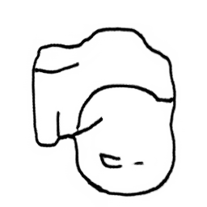}{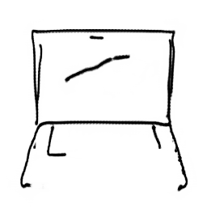}
        \srow{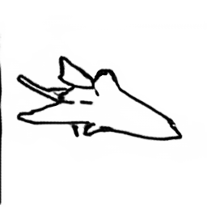}{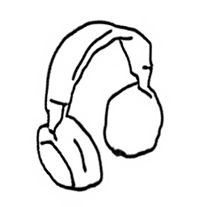}
        Input&
        Canny&
        HED&
        Contour&
        Ours&
        Input&
        Canny&
        HED&
        Contour&
        Ours\\
    }
}
\caption{Our results on photo-based sketch synthesis. \textbf{Top}: each sketch-photo pair: left: input photo, right: synthesized sketch. Results obtained on ShoeV2 and ChairV2.  \textbf{Bottom}: Results obtained on ShapeNet \cite{chang2015shapenet}. The column 1 is the input photo, Column 2-5 are lines generated by Canny, HED, Photo-Sketching\cite{li2019photo} (\textit{Contour} for short), and our model. Our model can generate line strokes with a hand-drawn effect, while HED and Canny detectors produce edge maps faithful to the original photos.  
Ours emphasize perceptually significant contours, not intensity-contrast significant as in edge maps.}
\label{fig:sketchsynthesis}
\end{figure}

\subsection{Photo-to-Sketch Synthesis}

\noindent\textbf{Synthesize a sketch given a photo.} 
As the shape translation network is bidirectional (i.e., $T$ and $T^\prime$), our model can also translate a photo into a sketch.  This task is not trivial, as human users can easily detect a fake sketch based on its stroke continuity and consistency. 
Fig.~\ref{fig:sketchsynthesis}(Top) shows that our generated sketches mimic manual line-drawings and emphasize contours that are perceptually significant.

\noindent\textbf{Sketch-like edge$+$ extraction.}
Sketch-to-photo and photo-to-sketch synthesis are opposite processes.  Synthesizing a photo based on a sketch requires the model to understand the structure of the object class and add information accordingly.  On the other hand, generating a sketch from a photo needs to throw away information, e.g., colors and textures.  This process may require less class prior to the opposite one.  Therefore, we suspect that our model can create sketches from photos in broader categories.

We test our shoe model directly on photos in ShapeNet \cite{chang2015shapenet}. Figure \ref{fig:sketchsynthesis}(Bottom) lists our results along with those from HED \cite{xie2015holistically} and Canny edge detector \cite{canny1986computational}. We also compare with Photo-Sketching \cite{li2019photo}, a method specifically designed for generating boundary-like drawing from photos. 
1) Unlike HED and Canny producing an edge map faithful to the photo, ours has a hand-drawn style, having learned the characteristics of sketches. 
2) Our model can dub as an edge$+$ extractor on unseen classes.  This is the most exciting corollary product: A promising automatic sketch generator that captures human visual perception beyond the edge map of a photo.

\begin{figure}[t]
\centering
\includegraphics[width=0.98\textwidth]{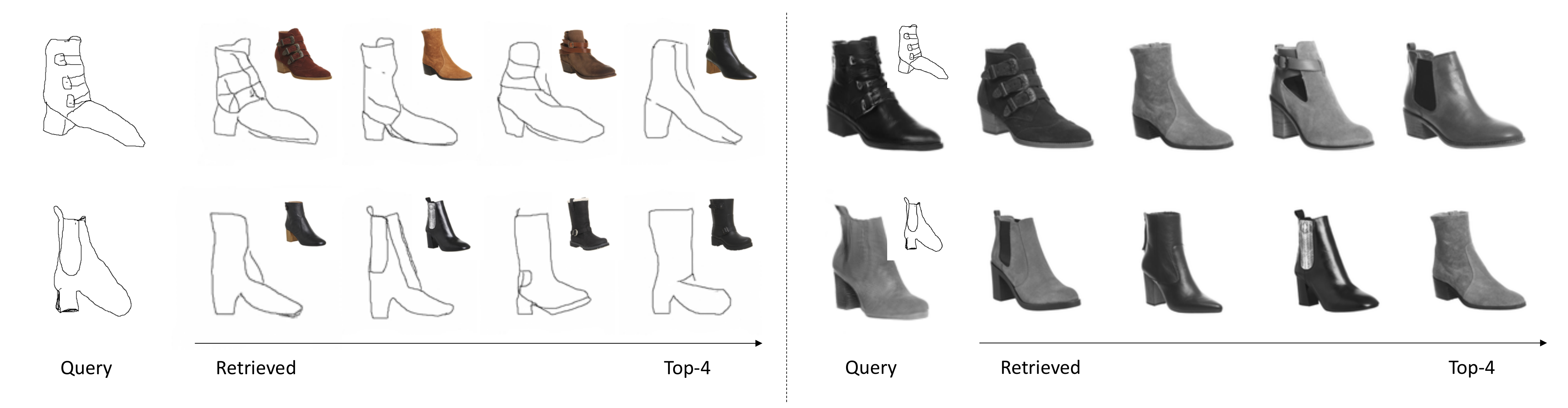}
\caption{Sample retrieval results (Top4). Our synthesis model can map photo to sketch domain and vice versa.  Cross-domain retrieval task can thus be converted to intra-domain retrieval. {\bf Left}: All candidate photos are mapped to sketches, thus both query and candidates are in the sketch domain. {\bf Right}: The query sketch is translated to a photo so that the query and candidates can be compared in the photo domain. Top right shows the original photo or sketch.}
\label{fig:re}
\end{figure}

\subsection{Application: Unsupervised Sketch-based Image Retrieval}

\noindent Sketch-based image retrieval (SBIR) is an important application of sketch. One of the main challenges faced by SBIR is the large domain gap. Common strategies include mapping sketches and photos into a common latent space or using edge maps as the intermediate representation.  However, our model enables direct mappings between these two domains. 

We thus conduct experiments in two possible mapping directions: {\bf 1)} Translate a sketch to a photo and then find its nearest neighbors in the photo gallery; {\bf 2)} Translate gallery photos to sketches, and then find the nearest sketches to the query sketch. Two pre-trained ResNet18 models \cite{he2016deep}, one on ImageNet while the other on TU-Berlin sketch dataset, are used as the feature extractors.

Figure \ref{fig:re} shows our retrieval results. Even \textit{without} any supervision, the results are already acceptable.  In the second experiment, we achieve an accuracy of 37.2\%(65.2\%) at top5 (top20) respectively. These results are higher than the results from \textit{sketch to edge map}, which are 34.5\%(57.7\%).

\section{Summary}
We propose an unsupervised sketch-to-photo synthesis model that can produce photos of high fidelity, realism, and diversity.  Our key insight is to decompose the task into geometrical shape translation and color content fill-in separately.  
Our model learns from a self-supervised denoising objective along with an attention, and allows diverse synthesis with an optional reference image.   
An exciting corollary product is a promising automatic sketch generator that captures human visual perception beyond the edgemap of a photo.

\bibliographystyle{splncs04}
\bibliography{egbib}

\section{Supplementary}

In the main paper, we propose a model for the task of sketch-based photo synthesis, which can deliver sketch-faithful realistic photos. Our key insight is to decompose this task into two separate translations.  Our two-stage model performs first geometrical shape translation in grayscale and then detailed content fill-in in color. Besides, at the first stage, a self-supervised learning objective along with noise sketch composition strategies and an attention module are brought up to handle abstraction and drawing style variations. 

In this Supplementary, we first provide further implementation details in Section \ref{sec:details}, including architectures of the proposed model, loss functions, and how we conducted the user study. Then in Section \ref{sec:qualitative}, we provide additional qualitative results to demonstrate the effectiveness of our model (see the caption of each figure for details). Additionally, we show results when applying our model in a \textit{multi-class} setting.

\subsection{Additional Implementation Details}\label{sec:details}
\noindent\textbf{Shape translation.} The architecture of two generators, $T$ and $T^\prime$, consists of nine residual blocks, two down-sampling, and two up-sampling layers. Instance normalization and ReLU is followed after each convolutional layer. The proposed \textit{attention module} includes two convolutional layers. We do not add a normalization layer after Conv layers in this module. The architecture of the \textit{discriminators}, $D_S$ and $D_G$, is composed of four convolutional layers, and each one is followed by instance normalization and LeakyReLU.

\noindent\textbf{Content enrichment.} The architecture of the \textit{encoder} $E$ consists of nine convolutional layers and two max-pooling layers, which shares the same structure of the first three blocks of VGG-19. The \textit{decoder} $D$ has twelve residual blocks and two up-sampling layers. When reference photos are available, $E$ is used for feature extraction. 

\noindent\textbf{Loss function.} For shape translation network, as indicated in the main paper, the loss function has five items: 
\begin{equation}
L_{adv}(T,D_{G};S,G) = \left(D_{G}(G)\right)^{2}+\left(1-D_{G}\left(T(S)\right)\right)^{2}
\end{equation}
\begin{equation}
L_{adv}(T^\prime,D_{S};G,S)) = \left(D_{S}(S)\right)^{2}+\left(1-D_{S}\left(T^\prime(G)\right)\right)^{2}
\end{equation}
\begin{equation}
 L_{cycle}(T, T^\prime;S,G)= \left\|S-T^\prime \left(T(S)\right)\right\|_{1} + \left\|G-T\left(T^\prime (G)\right)\right\|_{1}
\end{equation}
\begin{equation}
L_{identity}(T, T^\prime;S,G) = \left\|S-T^\prime(S)\right\|_{1} +  \left\|G-T(G)\right\|_{1}
\end{equation}
\begin{equation}
 L_{ss}(T, T^\prime;S,S^{noise})= \left\|S-T^\prime \left(T(S^{noise})\right)\right\|_{1}
\label{eq:ss}
\end{equation}
\noindent For content enrichment network $C$, the objective has four items, they are:
\begin{equation}
L_{adv}(C,D_{I};G,I) =\left(D_{I}(I)\right)^{2}+\left(1-D_{I}\left(C(G)\right)\right)^{2}
\end{equation}
\begin{equation}
L_{it}(C) = \left\|G-Lab\left(C\left(G \right)\right)\right\|_{1}
\end{equation}
\begin{equation}
 L_{cont}(C;G,R)= \left\|E(D(t))-t\right\|_{1}
\end{equation} 
\begin{equation}
\begin{aligned}
L_{style}(C;G,R) &= \sum_{i=1}^{K}\left\|\mu\left(\phi_{i}(D(t))\right)-\mu\left(\phi_{i}(R)\right)\right\|_{2} \\
&+ \sum_{i=1}^{K}\left\|\sigma\left(\phi_{i}(D(t))\right)-\sigma\left(\phi_{i}(R)\right)\right\|_{2}
\end{aligned}
\label{eq:style}
\end{equation}
\noindent where
\begin{equation}
t = AdaIN(E(G),E(R))
\end{equation}
\noindent $Lab(.)$ represents the conversion from RGB to Lab color space. $\phi_{i}(.)$ denotes a layer of a pre-trained VGG-19 model. In implementation, we use $relu1\_{1}$,
$relu2\_{1}$, $relu3\_{1}$, $relu4\_{1}$ layers with equal weights to compute style loss. 
The weights of these items, i.e., $\lambda_{1}$ to $\lambda_{7}$ in the main paper, are  1.0, 10.0, 0.5, 1.0, 10.0, 0.1 and 0.05 respectively.

\noindent\textbf{About user study.} One of the evaluation metrics we use is user study, i.e., \textit{Quality} (Table 1 in our main paper), it reflects how the generated photos are agreed with human imagination given a sketch. Specifically, for each comparison, an input sketch and its corresponding generated photos from two methods (one is the proposed method, and the other is a baseline method) are shown to a user at the same time, and then the user needs to choose which one is closer to his/her expectation. The range of the value is $[1, 100]$ while the default value for our method is set to 50. It is the ratio of cases that users prefer for the compared method. When a value is less than 50, it means that the generated photos of a baseline method are \textit{less} favored by volunteers compared with our proposed model; otherwise means people prefer the results of the baseline method.

\vspace{-0.3cm}
\subsection{Additional Qualitative Results}\label{sec:qualitative}
In this section, we provide more qualitative results (Figure \ref{fig:supp-2-sketchts} to Figure \ref{fig:supp_multiclass_shapenet}) to show the effectiveness of our model. 

\def\row#1#2#3#4{
\imw{supp_2_sketch2photo/input/#4_#1.png}{0.11}&
\imw{supp_2_sketch2photo/ours/#4_#1.png}{0.11}&
\imw{supp_2_sketch2photo/ref1/#4_#1.png}{0.11}&
\imw{supp_2_sketch2photo/input/#4_#2.png}{0.11}&
\imw{supp_2_sketch2photo/ours/#4_#2.png}{0.11}&
\imw{supp_2_sketch2photo/ref1/#4_#2.png}{0.11}&
\imw{supp_2_sketch2photo/input/#4_#3.png}{0.11}&
\imw{supp_2_sketch2photo/ours/#4_#3.png}{0.11}&
\imw{supp_2_sketch2photo/ref1/#4_#3.png}{0.11}\\
}
\def\prow#1#2{
\imw{supp-3-sketchy_tuberlin/sketchy_input/sketchy_#1.png}{0.12}&
\imw{supp-3-sketchy_tuberlin/sketchy_gray/sketchy_#1.png}{0.12}&
\imw{supp-3-sketchy_tuberlin/sketchy_RGB/sketchy_#1.png}{0.12}&
\imw{supp-3-sketchy_tuberlin/sketchy_ref/sketchy_#1.png}{0.12}&
\imw{supp-3-sketchy_tuberlin/tuberlin_input/tuberlin_#2.png}{0.12}&
\imw{supp-3-sketchy_tuberlin/tuberlin_gray/tuberlin_#2.png}{0.12}&
\imw{supp-3-sketchy_tuberlin/tuberlin_RGB/tuberlin_#2.png}{0.12}&
\imw{supp-3-sketchy_tuberlin/tuberlin_ref/tuberlin_#2.png}{0.12}\\
}
\begin{figure*}\centering
{\small
    \tb{@{}ccc|ccc|ccc@{}}{0.05}{
    \row{11}{7}{9}{shoes_supp}
    \row{8}{1}{2}{shoes_supp}
    \row{12}{13}{14}{shoes_supp}
    \row{15}{16}{17}{shoes_supp}
    \row{3}{9}{10}{chairs_supp}
    \row{6}{11}{8}{chairs_supp}
    Input&
    RGB &
    with Ref. &
    Input&
    RGB &
    with Ref. &
    Input&
    RGB &
    with Ref. \\
    }
}
\vspace{0.2cm}
{\scriptsize
    \tb{@{}cccc|cccc@{}}{0.05}{
    \prow{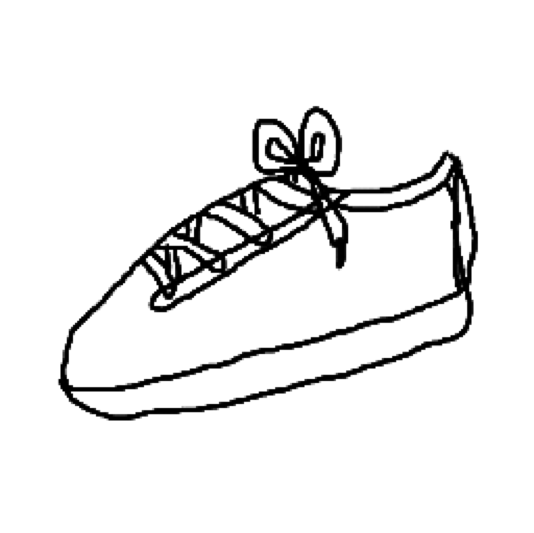}{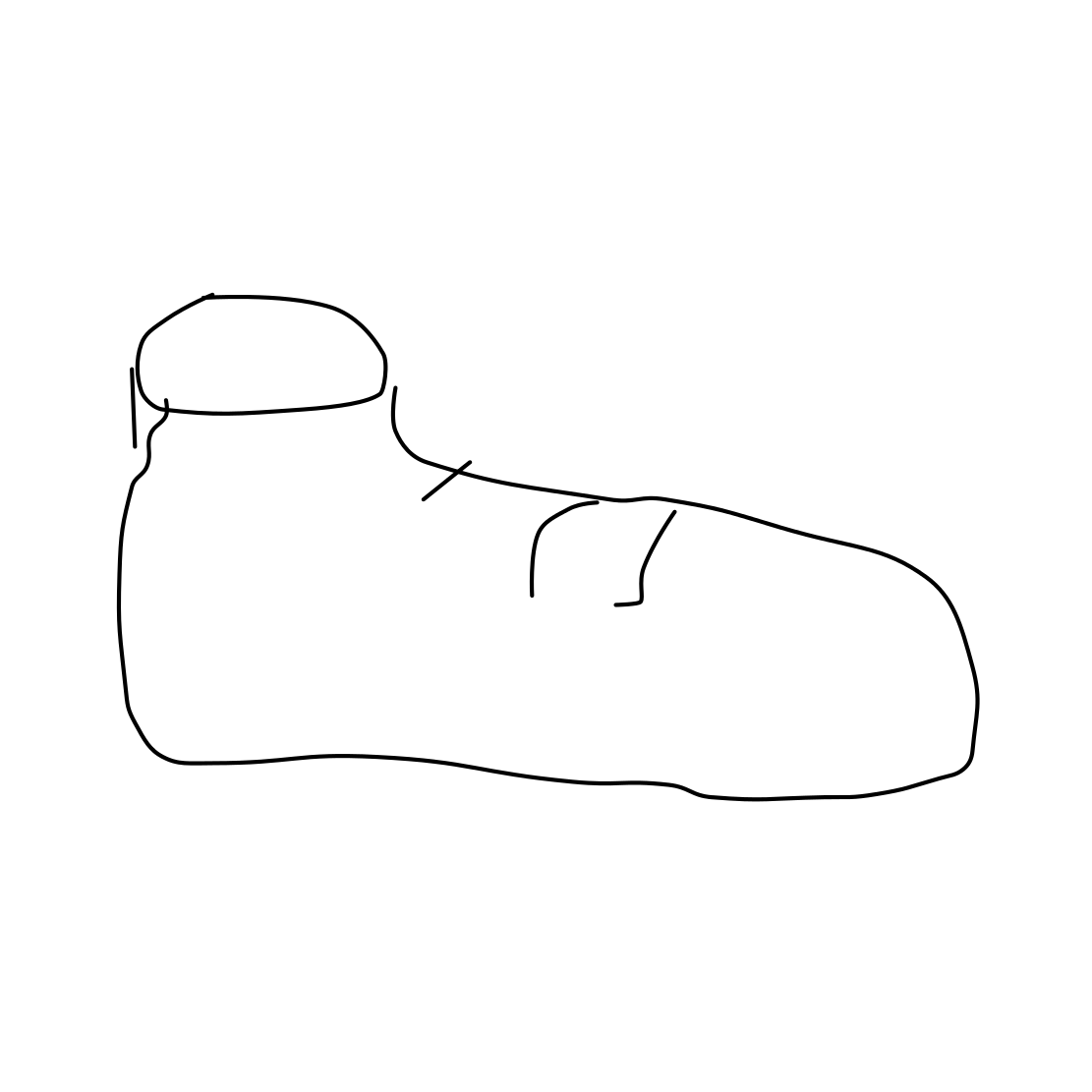}
    \prow{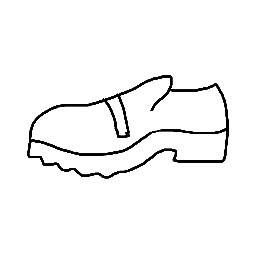}{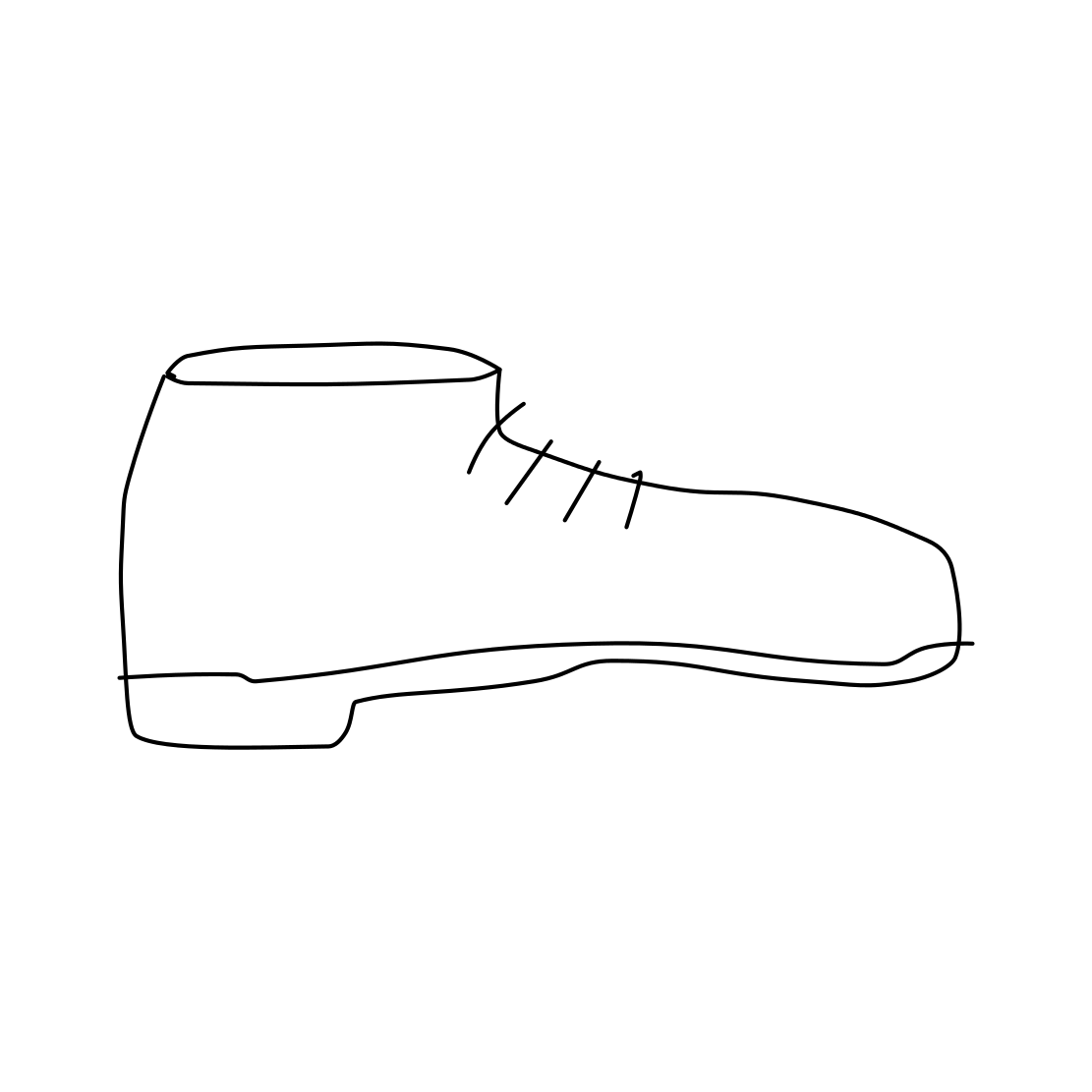}
    \prow{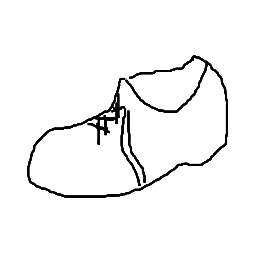}{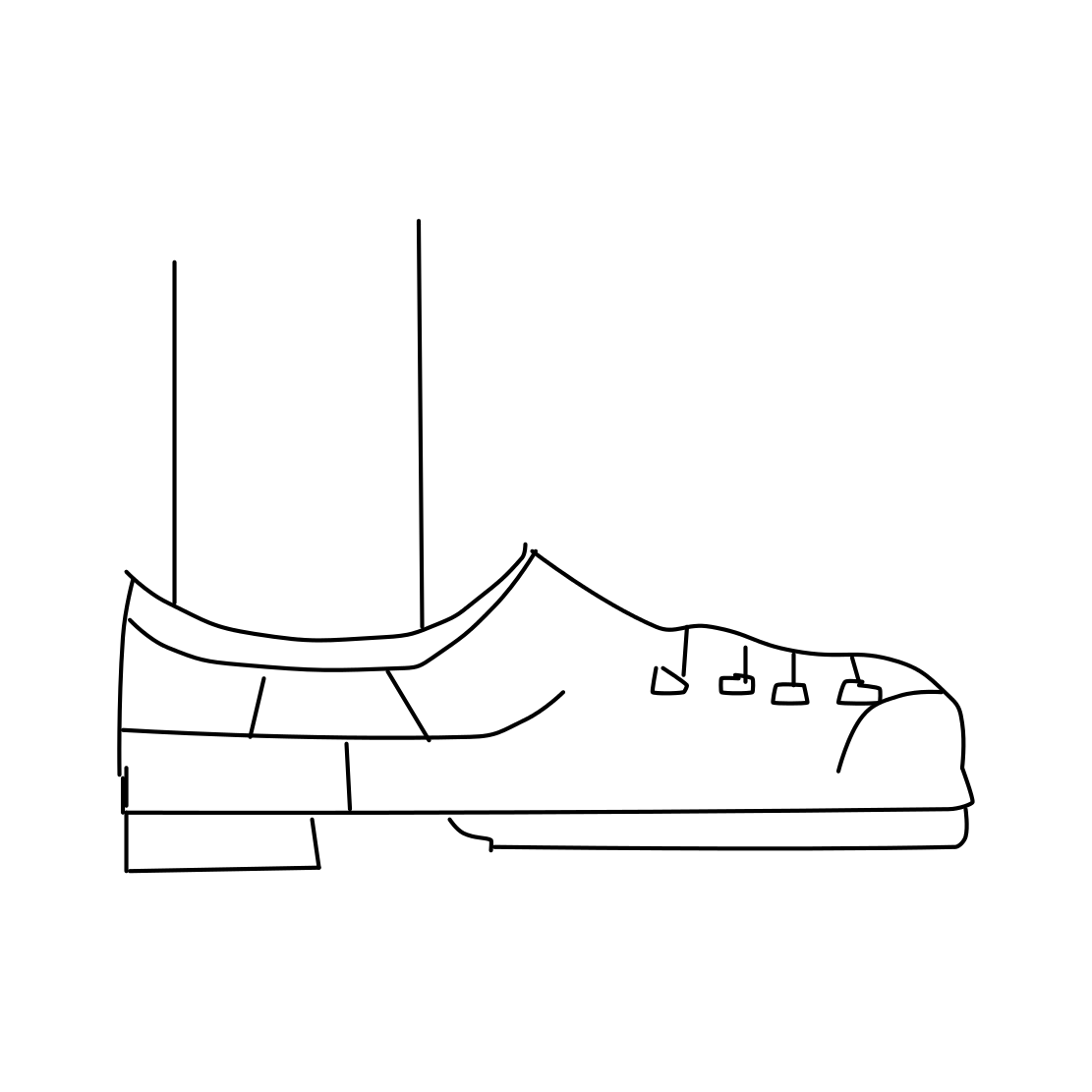}
    \prow{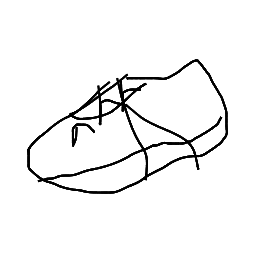}{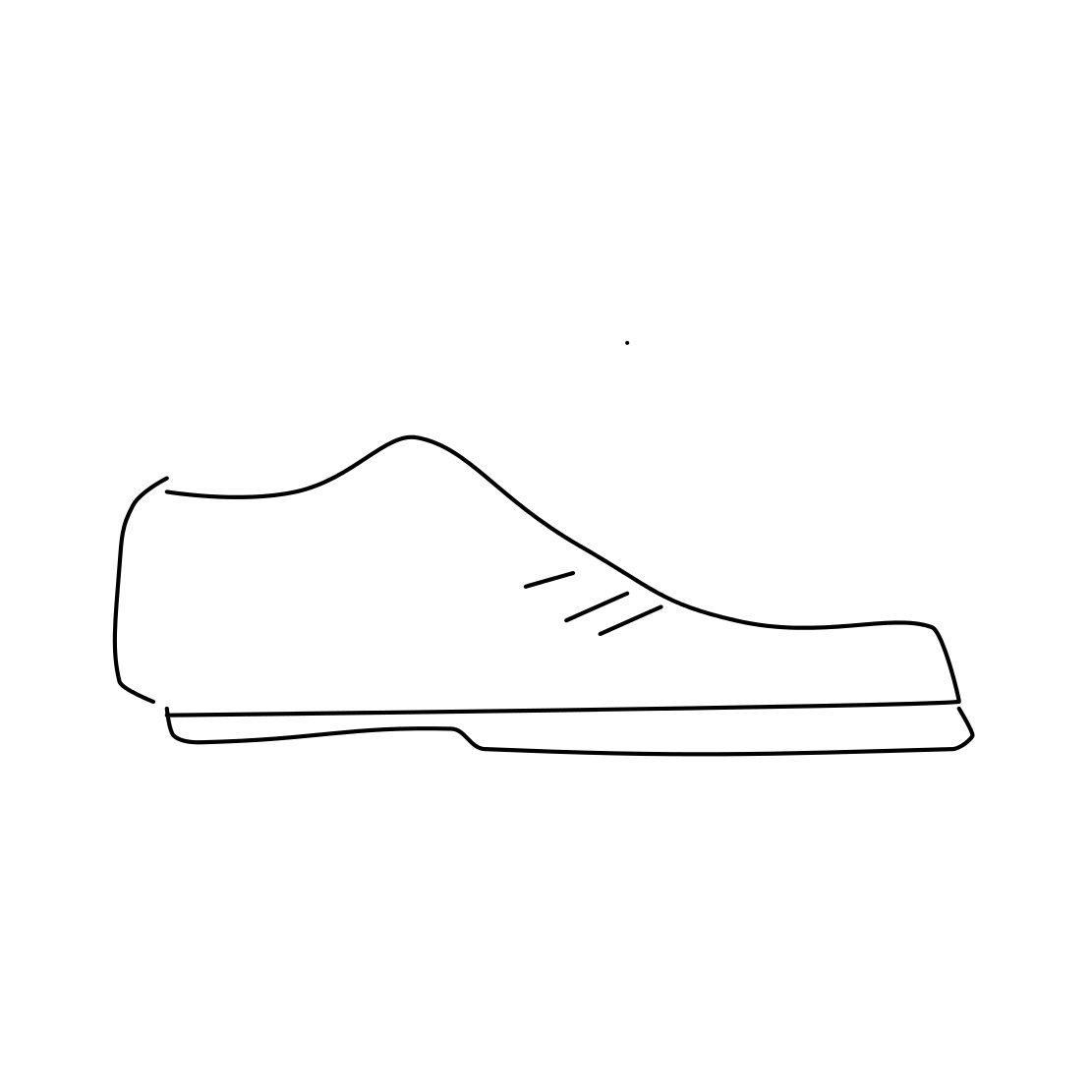}
    Input&
    Grayscale &
    RGB &
    with Ref. &
    Input&
    Grayscale &
    RGB &
    with Ref. \\
    }
}
\vspace{-0.2cm}

\caption{\textbf{Top:} Synthesized results obtained by our model, with (the 3rd column) and without (the 2nd column) references. Reference images are shown in the top right corner. \textbf{Bottom:} Generalization across sketch datasets. On the left, input sketches are from Sketchy dataset; while sketches on the right side are from TU-Berlin dataset. The rest images, from left to right, are synthesized grayscale images, synthesized RGB photos, and RGB photos when reference images are available. Note that all these results are produced by our model trained on ShoeV2 dataset.}
\label{fig:supp-2-sketchts}
\end{figure*}

\def\row#1#2{
\imw{supp_4_attention/complex_sketch/#1.png}{0.10}&
\imw{supp_4_attention/attention/#1.png}{0.10}&
\imw{supp_4_attention/ours_rec/#1.png}{0.10}&
\imw{supp_4_attention/ours_RGB/#1.png}{0.10}&
\imw{supp_4_attention/ugatit/#1.png}{0.10}&
\imw{supp_4_attention/complex_sketch/#2.png}{0.10}&
\imw{supp_4_attention/attention/#2.png}{0.10}&
\imw{supp_4_attention/ours_rec/#2.png}{0.10}&
\imw{supp_4_attention/ours_RGB/#2.png}{0.10}&
\imw{supp_4_attention/ugatit/#2.png}{0.10}\\
}
\begin{figure*}[t]\centering
{\small
\tb{@{}ccccc|ccccc@{}}{0.05}{
\row{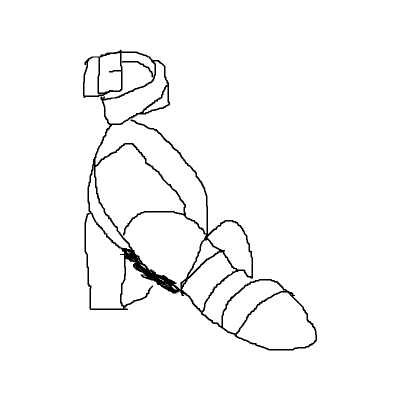}{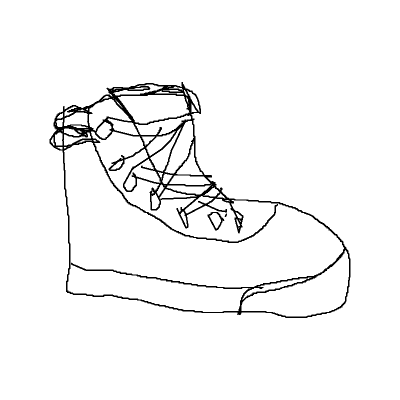}
\row{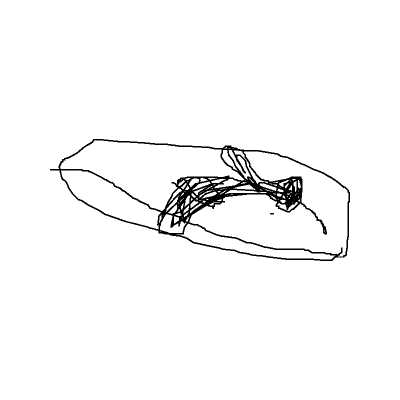}{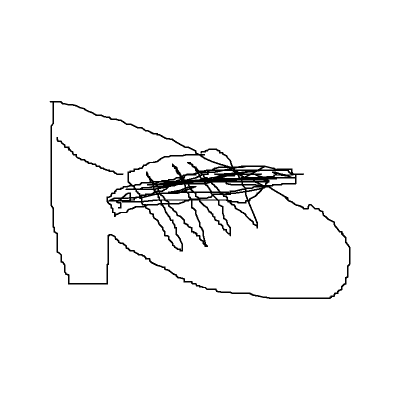}
\row{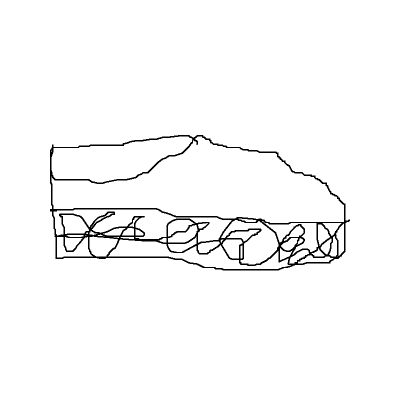}{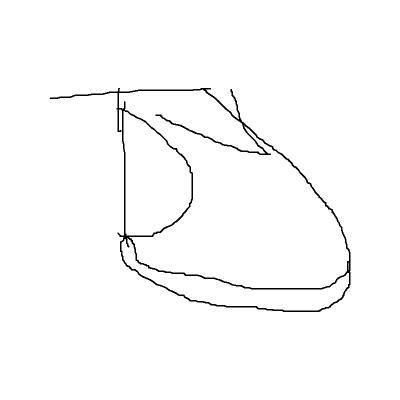}
(a)&
(b)&
(c)&
\textbf{(d)}&
(e)&
(a)&
(b)&
(c)&
\textbf{(d)}&
(e)\\
}
}
\caption{Our model can deal with noise sketches. (a) are input sketches; (b)(c)(d) show learned attention masks,  \textit{reconstructed} sketches, and photos synthesized by our model. (e) are the results of UGATIT. It is clear to see that our model can handle noise sketches better than UGATIT. Besides, the disparity between (a) and (c) indicates what irrelevant noise strokes are \textit{ignored} by our model.}
\label{fig:supp-4-attention}
\end{figure*}

\def\row#1#2#3#4{
\imw{supp_5_photo2sketch/shoes/#1.png}{0.12}&
\imw{supp_5_photo2sketch/sketches/#1.png}{0.12}&
\imw{supp_5_photo2sketch/shoes/#2.png}{0.12}&
\imw{supp_5_photo2sketch/sketches/#2.png}{0.12}&
\imw{supp_5_photo2sketch/shoes/#3.png}{0.12}&
\imw{supp_5_photo2sketch/sketches/#3.png}{0.12}&
\imw{supp_5_photo2sketch/shoes/#4.png}{0.12}&
\imw{supp_5_photo2sketch/sketches/#4.png}{0.12}\\
}
\def\chairrow#1#2#3#4{
\imw{supp_5_photo2sketch/chairs/#1.png}{0.12}&
\imw{supp_5_photo2sketch/chairs_sketches/#1.png}{0.12}&
\imw{supp_5_photo2sketch/chairs/#2.png}{0.12}&
\imw{supp_5_photo2sketch/chairs_sketches/#2.png}{0.12}&
\imw{supp_5_photo2sketch/chairs/#3.png}{0.12}&
\imw{supp_5_photo2sketch/chairs_sketches/#3.png}{0.12}&
\imw{supp_5_photo2sketch/chairs/#4.png}{0.12}&
\imw{supp_5_photo2sketch/chairs_sketches/#4.png}{0.12}\\
}
\begin{figure*}[t]\centering
{\small
\tb{@{}cccccccc@{}}{0.05}{
\row{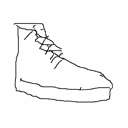}{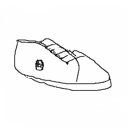}{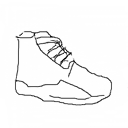}{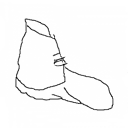}
\row{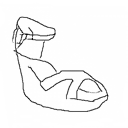}{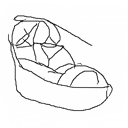}{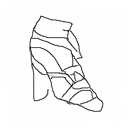}{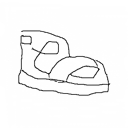}
\row{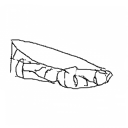}{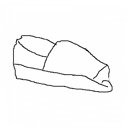}{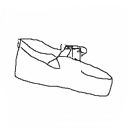}{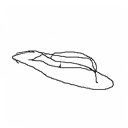}
\row{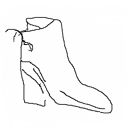}{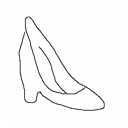}{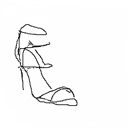}{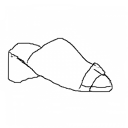}
\chairrow{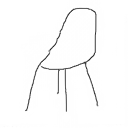}{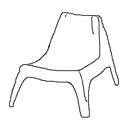}{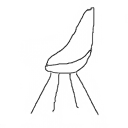}{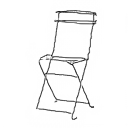}
\chairrow{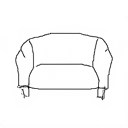}{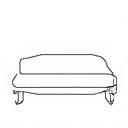}{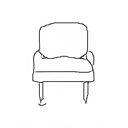}{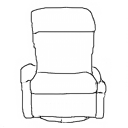}
(a)&
(b)&
(a)&
(b)&
(a)&
(b)&
(a)&
(b)\\
}
}
\caption{Results of photo-based sketch synthesis. (a) Input photo, (b) synthesized sketch by our model. The synthesized sketches can not only reflect the distinguishing feature of original objects, but also mimic different drawing styles. For example, in the first row, \textit{shoelace} are depicted in different styles.}
\label{fig:supp-5-photo2sketch}
\end{figure*}

\def\row#1#2{
\imw{supp_6_shapenet/photos/#1.png}{0.1}&
\imw{supp_6_shapenet/canny/#1.png}{0.1}&
\imw{supp_6_shapenet/HED/#1.png}{0.1}&
\imw{supp_6_shapenet/photo_sketching/#1.png}{0.1}&
\imw{supp_6_shapenet/ours/#1.png}{0.1}&
\imw{supp_6_shapenet/photos/#2.png}{0.1}&
\imw{supp_6_shapenet/canny/#2.png}{0.1}&
\imw{supp_6_shapenet/HED/#2.png}{0.1}&
\imw{supp_6_shapenet/photo_sketching/#2.png}{0.1}&
\imw{supp_6_shapenet/ours/#2.png}{0.1}\\
}
\def\rrow#1#2{
\imw{supp_6_shapenet/canny/p#11.png}{0.1}&
\imw{supp_6_shapenet/photos/p#12.png}{0.1}&
\imw{supp_6_shapenet/HED/p#13.png}{0.1}&
\imw{supp_6_shapenet/photo_sketching/p#12.png}{0.1}&
\imw{supp_6_shapenet/ours/p#14.png}{0.1}&
\imw{supp_6_shapenet/canny/p#21.png}{0.1}&
\imw{supp_6_shapenet/photos/p#22.png}{0.1}&
\imw{supp_6_shapenet/HED/p#23.png}{0.1}&
\imw{supp_6_shapenet/photo_sketching/p#22.png}{0.1}&
\imw{supp_6_shapenet/ours/p#24.png}{0.1}\\
}
\begin{figure*}[t]\centering
{\small
\tb{@{}ccccc|ccccc@{}}{0.05}{
\row{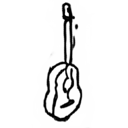}{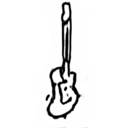}
\row{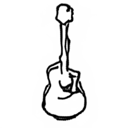}{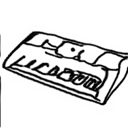}
\row{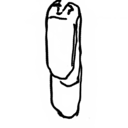}{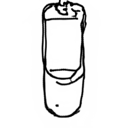}
\row{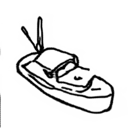}{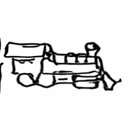}
\row{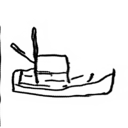}{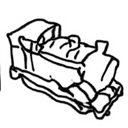}
\row{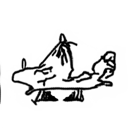}{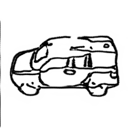}
\row{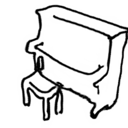}{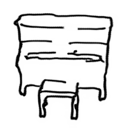}
\row{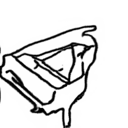}{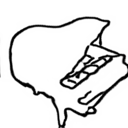}
(a)&
(b)&
(c)&
(d)&
(e)&
(a)&
(b)&
(c)&
(d)&
(e)\\
}
}
\caption{Results obtained on ShapeNet \cite{chang2015shapenet}. (a) are input photos, (b) to (e) are lines derived by Canny \cite{canny1986computational}, HED \cite{xie2015holistically}, Photo-Sketching \cite{li2019photo}, and our shoe model. Our model can generate lines with a hand-drawn effect, while HED and Canny detectors produce edgemaps faithful to original photos. Comparing with results of Photo-Sketching, ours are visually more similar with free-hand sketches. }
\label{fig:supp-6-shapenet}
\end{figure*}

\def\row#1#2#3#4{
\imw{supp_multiclass_shapenet/sketch/#1.png}{0.12}&
\imw{supp_multiclass_shapenet/photo/#1.png}{0.12}&
\imw{supp_multiclass_shapenet/sketch/#2.png}{0.12}&
\imw{supp_multiclass_shapenet/photo/#2.png}{0.12}&
\imw{supp_multiclass_shapenet/sketch/#3.png}{0.12}&
\imw{supp_multiclass_shapenet/photo/#3.png}{0.12}&
\imw{supp_multiclass_shapenet/sketch/#4.png}{0.12}&
\imw{supp_multiclass_shapenet/photo/#4.png}{0.12}\\
}
\begin{figure*}[t]\centering
{\small
    \tb{@{}cccccccc@{}}{0.05}{
    \row{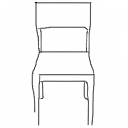}{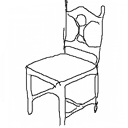}{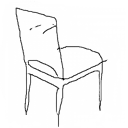}{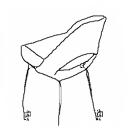}
    \row{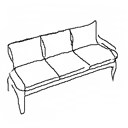}{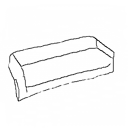}{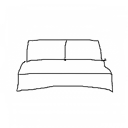}{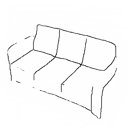}
    \row{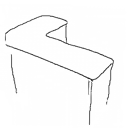}{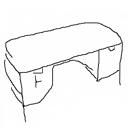}{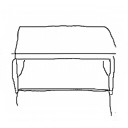}{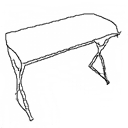}
    \row{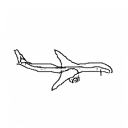}{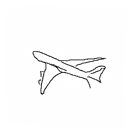}{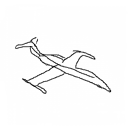}{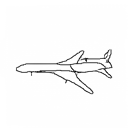}
    \row{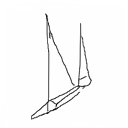}{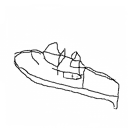}{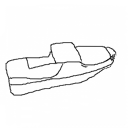}{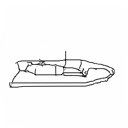}
    \row{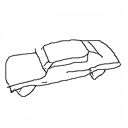}{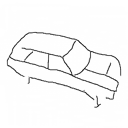}{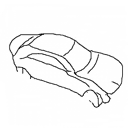}{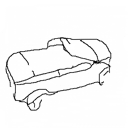}
    (a)&
    (b)&
    (a)&
    (b)&
    (a)&
    (b)&
    (a)&
    (b)\\
    }
}
\caption{Results of \textbf{multi-class sketch-to-photo synthesis} on ShapeNet dataset. Given performance achieved in the single-class setting, we wonder if our proposed model can work for multi-classes. We thus conduct experiments on ShapeNet. To be specific, we select 11 classes, each contains photos varying from 300 to 8000, and form training and testing set with 20,656 and 5,823 photos respectively. Then we generate fake \textit{sketches} using our shoe model. Next, we train our \textit{shape translation} network on the newly formed multi-class image set. All training settings are the same as training in a single class, and class information is \textbf{not} used during training. Resutls are displayed above. (a) are input sketches, and (b) are synthesized grayscale images. Examples in each row are from the same class. To our surprise, the model can generate photos for multiple classes, even \textbf{without} any class information. We assume that our model is capable of gaining semantic understanding during the class-agnostic training process. }
\label{fig:supp_multiclass_shapenet}
\end{figure*}

\end{document}